\documentclass{article}

\def\arxivpreprint{}

\ifdefined\pdfminorversion
\fi

\usepackage{iclr2027_conference}
\usepackage{times}

\usepackage{amsmath,amsfonts,bm}

\def\1{\bm{1}}

\DeclareMathAlphabet{\mathsfit}{\encodingdefault}{\sfdefault}{m}{sl}
\SetMathAlphabet{\mathsfit}{bold}{\encodingdefault}{\sfdefault}{bx}{n}

\usepackage{amsmath}
\usepackage{amssymb}
\usepackage{booktabs}
\usepackage{flafter}
\usepackage{graphicx}
\usepackage{xcolor}
\definecolor{uoftblue}{HTML}{002A5C}
\usepackage{hyperref}
\hypersetup{colorlinks=true, allcolors=uoftblue}
\usepackage{url}
\usepackage[nameinlink,noabbrev]{cleveref}

\newcommand{\benchmarkname}{\textsc{ArbiGraph}}
\newcommand{\benchmarktitle}{%
\benchmarkname{}:\\
Composing Arbitrary Verifiable Tasks\\for Agent Evaluation%
}

\title{\benchmarktitle}

\newcommand{\paperauthors}{%
Pavel Golikov\textsuperscript{1,2},
Evgenii Opryshko\textsuperscript{1,2},
Gennady Pekhimenko\textsuperscript{1,2},
Mark C. Jeffrey\textsuperscript{1}%
}
\newcommand{\paperaffiliations}{%
\textsuperscript{1}University of Toronto,
\textsuperscript{2}Vector Institute%
}

\author{%
\makebox[\dimexpr\textwidth-2\tabcolsep\relax][c]{\paperauthors}\\[-0.15em]
\makebox[\dimexpr\textwidth-2\tabcolsep\relax][c]{\normalfont\paperaffiliations}%
}

\ifdefined\arxivpreprint
\iclrfinalcopy
\fi
\begin{document}

\raggedbottom

\maketitle
\ifdefined\arxivpreprint
\lhead{}
\renewcommand{\headrulewidth}{0pt}
\fi

\begin{abstract}
Language model agents must often solve a sequence of problems and retain earlier answers as
the workflow grows. Evaluating problems one at a time does not show how well agents retain
and use earlier answers.
To evaluate this ability, we introduce \benchmarkname{}, a framework for generating benchmark instances that combine problems with verifiable answers
into graphs of dependent tasks. Each task is a problem, and each connection specifies an
earlier answer that the agent must report to obtain the next problem's input.
An incorrect report returns an alternative input whose verified answer differs from the
intended answer. This mechanism supports tasks from different domains without requiring
conversions between their input and output types.
We test the framework with math problems and tasks that ask agents to predict the output
of Python code. We evaluate three models on the same problems both in isolation and in
three topologies, each containing 18 tasks. We construct an independent-error baseline
by multiplying the isolated accuracies of the final task and all tasks it depends on.
Averaged across the three topologies and the math and Python categories, measured accuracy falls below this baseline
by 45.0 percentage points for Gemma-4-31B-it, 35.6 for Qwen3.8-27B, and 11.0 for GPT-OSS-120B.
These additional losses indicate that solving problems within a larger workflow adds difficulty beyond
the accumulation of errors on isolated problems. We also study how graph structure affects
accuracy on the final task while keeping the number of tasks fixed.
\ificlrfinal
ArbiGraph is available at \url{https://github.com/pavelgolikov/ArbiGraph.git}
\fi


\end{abstract}

\section{Introduction}

Language model agents use tools to solve problems over many steps.
As a workflow grows, agents must keep track of earlier answers, update them when needed,
and ignore information that is no longer useful. Solving each problem correctly in isolation
is not enough to ensure success across the whole workflow.
Existing evaluations show that models struggle with distracting information, changes in the
position of relevant information, and long interaction histories~\citep{liu2024lostinthemiddle,hsieh2024ruler,bai2024longbench,bai2024mtbench101,maharana2024locomo}.
These evaluations examine how models use context through documents, conversations, and memory tasks.
Other work combines multiple tasks: REST presents several problems at once~\citep{pan2025rest},
SIFo tests sequences of instructions~\citep{chen2024sifo}, and Composition-RL combines verifiable
problems into new questions~\citep{xu2026compositionrl}.
RACES connects tasks with compatible input and output types using a set of composition
operators~\citep{xiang2026races}.
We study how agents perform when the same problems are connected in different ways.
This lets us vary the dependencies while keeping the difficulty of the individual problems fixed.

\begin{figure}[t]
    \centering
    \includegraphics[width=\linewidth]{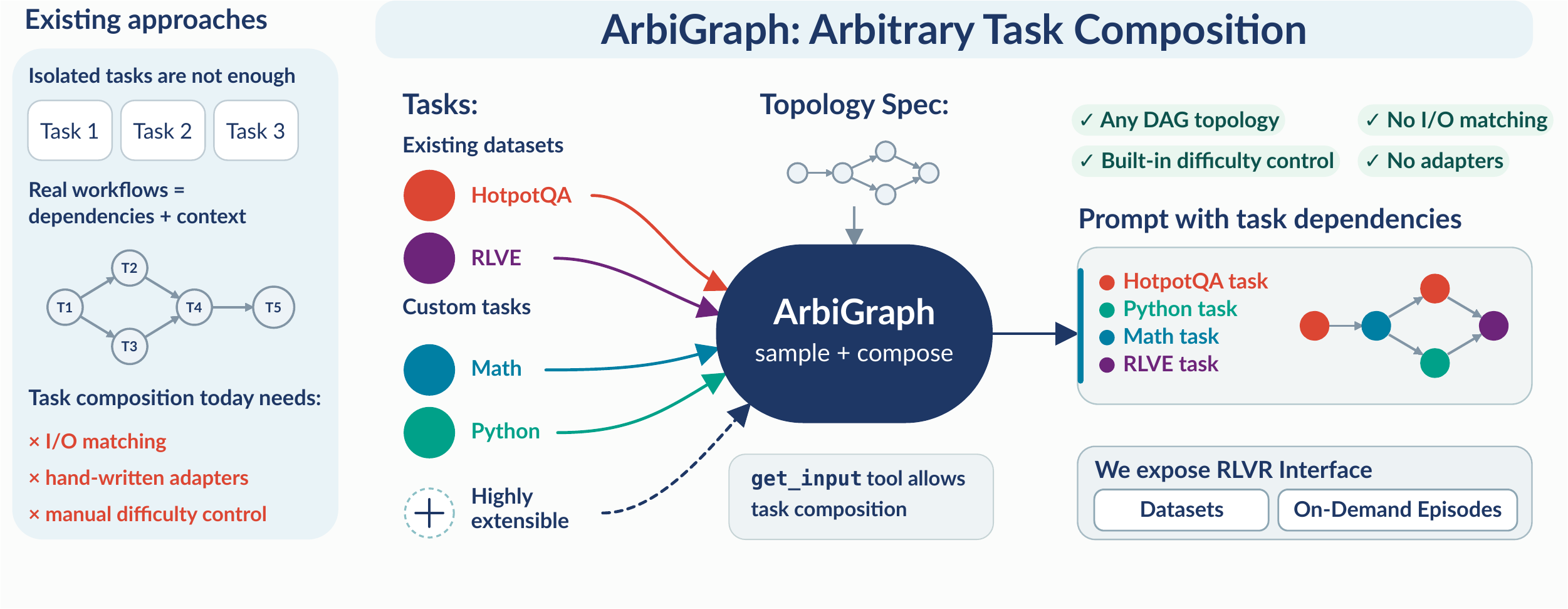}
    \caption{Overview of \benchmarkname{}. Custom tasks and existing datasets provide native,
    independently verifiable sources for user-specified task graphs, without pairwise adapters
    or input/output type matching. Answer-gated input access supports agent evaluation and RLVR
    training.}
    \label{fig:overview}
\end{figure}


Connecting existing problems is difficult when one problem's answer must become another's input.
Their input and output types may differ, requiring a conversion and instructions that explain
it to the agent. A task that depends on several earlier answers also needs a rule for combining
them. Some conversions can hide errors: rounding, for example, can turn both a correct and an
incorrect answer into the same input for the next task.

We introduce \benchmarkname{}, a framework for generating benchmark instances from problems
with verifiable answers. Each instance is a graph: every node contains a concrete problem, and
each edge identifies an earlier answer needed by a later task. Each problem keeps its original input format and
answer format.
To request a task's input, the agent reports the answers to the tasks it depends on.
A tool, \texttt{get\_input}, checks these reports and returns the intended input if they are correct.
Otherwise, it returns an alternative input with a different verified answer.
A second tool, \texttt{submit\_answers}, accepts the agent's answer to the designated final task for grading.
Because earlier answers are checked before an input is supplied, tasks from different domains
can be connected without converting their answers into one another's input types.

The framework supports different kinds of tasks with verifiable answers.
Our implementation includes math problems, predicting the output of Python code, algorithmic
problems from Reinforcement Learning with Adaptive Verifiable Environments (RLVE)~\citep{zeng2026rlve},
and multi-hop questions from HotpotQA's distractor setting~\citep{yang2018hotpotqa}.
Adding a task category requires examples or a generator, along with a way to check answers
and provide alternative inputs with different answers.
The workflows can also support reinforcement learning with verifiable rewards (RLVR), with
a reward determined by the final answer. We provide this interface but do not train models in
this work. The framework requires agents that can call tools.

We evaluate three models on individual problems and on instances containing 18 math or 18 Python
tasks. We construct an independent-error baseline by multiplying the isolated accuracies of the
final task and all tasks it depends on. Accuracy on the final task falls below this baseline,
with the additional loss varying across models.
We also study changes in topology, task order, and task categories.
For the last comparison, we replace half the tasks in each math-only instance with tasks from other
categories. After accounting for differences in the difficulty of the individual problems,
GPT-OSS-120B loses accuracy. For Qwen3.8-27B and Gemma-4-31B-it, the uncertainty intervals include zero.
A branch-pruning experiment further shows that agents sometimes correctly solve tasks
on a branch disconnected from the requested target.


Our main contributions are:
\begin{itemize}
    \item A method for composing problems with verifiable answers. It checks earlier answers
    to determine the next task's input, so incorrect reports lead to inputs with different answers.
    No conversion between task input and output types is needed.
    \item \benchmarkname{}, a framework that generates benchmark instances with chosen
    topologies and task pools, with tools for evaluation and training with verifiable rewards.
    \item An evaluation of problems in isolation and in workflows, including workflows that
    mix task categories or contain branches unnecessary for the target. We measure how much
    accuracy falls below the independent-error baseline and whether agents skip unnecessary work.
\end{itemize}

\begin{figure}[t]
    \vspace{-4mm}
    \centering
    \includegraphics[width=\linewidth]{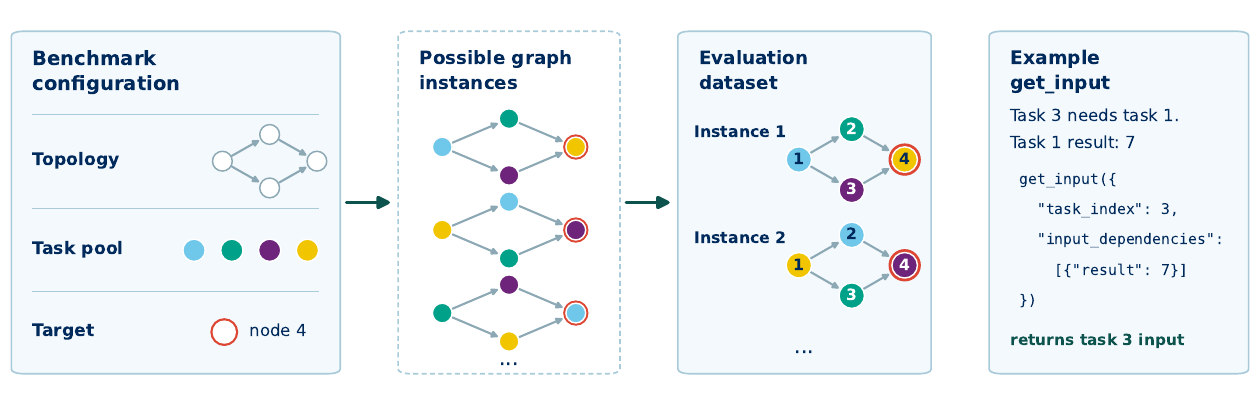}
    \caption{Framework terms and an example tool call. On the left, a benchmark configuration
    determines the possible graph instances. Sampling from it produces an evaluation dataset. Four colors
    distinguish example tasks, and the red outline marks the target. On the right, task 3
    depends on task 1. The agent passes task 1's reported answer in
    \texttt{input\_dependencies}. The tool checks it before returning an input for task 3.}
    \label{fig:framework-terms}
    \vspace{-2mm}
\end{figure}

\section{The \benchmarkname{} Framework}
\label{sec:design}

\benchmarkname{} connects tasks through tool calls. Each generated \emph{instance}
is a graph with one task at each node. When the framework creates an instance, it
fixes one intended input and its verified answer for each task. To request a node's input, the
agent reports answers to its prerequisites via a tool call. The tool returns the intended input if all reports
are correct; otherwise, it returns another valid input with a different verified answer.
This lets us check intermediate answers while connected tasks keep their own input
and answer formats.

The instance's \emph{topology} specifies its nodes and dependency edges. An edge points from a
prerequisite task to a task that needs its answer. One node is the \emph{target}. The
agent's answer to the target determines the score for the instance. The framework also keeps
reported answers at other nodes so that earlier errors can be studied.

A \emph{task pool} contains the tasks available for assignment to nodes. A
\emph{benchmark configuration} specifies the topology, task pool, target, and how tasks are
sampled. Together, these determine the \emph{possible graph instances}. An \emph{evaluation
dataset} contains concrete instances generated from that configuration.
\Cref{fig:framework-terms} shows how these terms fit together.

\subsection{Task Node}
We represent the task at node \(i\) as
\[
    T_i=(p_i,x_i,y_i,\mathcal B_i,c_i),
\]
where \(p_i\) is the problem statement, \(x_i\) is the \emph{intended input}, and
\(y_i\) is its verified answer. The \emph{alternative-input bank}
\(\mathcal B_i=\{(\tilde x_i^{(k)},\tilde y_i^{(k)})\}_k\) contains valid input--answer pairs
whose verified answers differ from \(y_i\) and from one another under \(c_i\). The \emph{answer
comparator} \(c_i\) determines whether answers match;
it can require exact equality or allow a numerical tolerance. Answers can be
verified by an executable solver or labels. Correctly executing the task on any bank input
\(\tilde x_i^{(k)}\) produces its verified answer \(\tilde y_i^{(k)}\), which the comparator rejects
against the intended answer: \(c_i(\tilde y_i^{(k)},y_i)=0\).
The agent sees \(p_i\) and the input returned
by the tool, but not the stored verified answers.

\subsection{Task Composition}
The \emph{parents} of node \(i\) are its direct prerequisites. We denote them by
\(\mathrm{pa}(i)\). To get the input for node \(i\), the agent calls \texttt{get\_input}. The tool call arguments specify the task index and a list with answers to the parent tasks. The prompt specifies the order of
the parents in this list. \Cref{fig:framework-terms} shows a call for a node with one parent.
We write the reported values as \(\hat{\mathbf y}_{\mathrm{pa}(i)}\). The tool checks
each value with the corresponding parent's answer comparator. It returns
\[
\text{input for node }i =
\begin{cases}
x_i, & \text{if }c_j(\hat y_j,y_j)=1\text{ for every }j\in\mathrm{pa}(i),\\
\tilde x_i^{(k)}, & \text{otherwise, with }(\tilde x_i^{(k)},\tilde y_i^{(k)})\in\mathcal B_i.
\end{cases}
\]
Root nodes have no parents, but they still require a tool call with an empty list of answers to receive their
intended input. This is also true for our single-task evaluations, which makes them directly comparable with larger
topologies.
The agent submits its target answer through a \texttt{submit\_answers} tool.

The user provides a JSON file that lists the nodes, edges, problem types, and target. The
framework uses NetworkX~\citep{hagberg2008networkx} to check that the dependencies
contain no cycles, then assigns problems to
nodes from prerequisites to dependents. It also supports visualizations using
Graphviz~\citep{ellson2004graphviz}. The topology can vary the number and arrangement of
dependent or irrelevant problems without rewriting the problems. Any task in the topology can be
selected as the target.

\subsection{Why Direct Adapters Are Insufficient}
\label{sec:adapters}
Composing problems by passing outputs as inputs requires compatible formats or an adapter to convert between them.
A node with several parents also needs a rule for combining
their answers. The prompt must explain each conversion to the agent, and that explanation
must match the code. These requirements limit which problems can be connected and add
instructions unrelated to the problems.
Direct composition also makes a downstream task's input depend on upstream outputs, which
can lead to difficulty accumulation. Controlling difficulty may therefore require additional constraints
on these outputs or their conversion. For example, in math, Python, and other technical tasks,
numeric magnitudes or sequence lengths can grow unchecked across successive tasks, requiring
explicit bounds or truncation to keep downstream inputs manageable.
Additionally, we observed that conversions could hide errors.
For example, rounding may map a wrong parent answer to the
same child input as the correct answer. This problem was especially prevalent when the child task had few possible inputs. Even without an
explicit conversion, two different inputs may have the same answer. Checking the reported
answers and selecting inputs that result in distinct answers avoids both cases.

\subsection{Task Categories}
\benchmarkname{} currently supports math, Python tracing, RLVE algorithmic tasks, and
multi-hop reading comprehension in HotpotQA's distractor setting. Adding a category requires
examples or a generator, an answer comparator, and alternative inputs with distinct answers.

\noindent\textbf{Math.}\enspace The category consists of 40 operators selected by hand.
They cover four input-output formats: 10 list-to-list tasks,
10 list-to-scalar tasks, 10 scalar-to-scalar tasks, and 10 scalar-to-list tasks.
The tasks are drawn from areas that are algorithmically concrete and exactly computable: linear
algebra, polynomial arithmetic, discrete transforms, combinatorics, geometry, and number theory.
Python solvers verify the answer for each sampled input.

\par\noindent\textbf{Python tracing.}\enspace The category contains 80 function/input variants drawn from the LeetCode
Dataset~\citep{leetcode_dataset}.
The agent traces a concrete function on concrete inputs and reports the returned value without using tools to run the code. These functions include array and list processing, sorting and searching, stack and greedy algorithms,
dynamic programming, simulation, and arithmetic routines.

\par\noindent\textbf{RLVE.}\enspace The category adapts 40 algorithmic objectives from RLVE-Gym~\citep{zeng2026rlve}.
Their underlying problems are retained, but the implementations are integrated with
\benchmarkname{}'s input sampling, executable solvers, and alternative-input banks.

\begin{figure}[t]
    \vspace{-2mm}
    \centering
    \includegraphics[width=\linewidth]{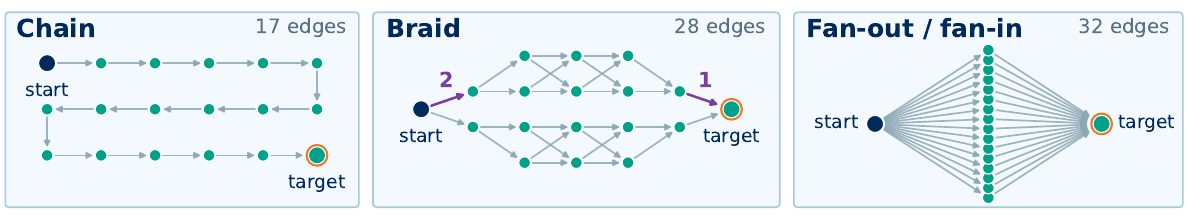}
    \caption{The three topologies with 18 nodes used in \Cref{sec:experiments}. Every node other than the target is an ancestor of the target. The \emph{shuffled
    braid} (not shown) keeps the braid's problems and dependencies but reassigns task numbers.
    Edges~1 and~2 in the braid panel mark the cuts used for branch pruning.}
    \label{fig:topologies}
    \vspace{-2mm}
\end{figure}

\par\noindent\textbf{HotpotQA (distractor setting).}\enspace The category draws 200 distinct labeled
questions from the HotpotQA training data in distractor format~\citep{yang2018hotpotqa}.
Each input supplies the source question and all ten context paragraphs, including distractors,
with paragraph titles and sentence boundaries preserved. The agent returns a short answer and
supporting-fact locations in a structured result.

\subsection{RLVR Interface}
The same instances can be used for reinforcement learning with verifiable rewards
(RLVR). An instance is sampled from a benchmark configuration or loaded from a saved dataset.
For each agent attempt, the agent receives the instance's prompt and tool definitions.
An accepted target answer earns a reward of one if it matches the target's verified answer,
and zero otherwise.
We record correctness at every node for diagnostics; future work could also use these records for reward shaping or advantage attribution.



\section{Methodology}

\label{sec:methodology}
We compare an agent's accuracy on individual problems with its accuracy on the same problems
inside instances with 18 nodes. Each problem keeps its statement, parameters, intended input, verified
answer, and alternative-input bank in every instance that contains it, including instances where it
is the only node. All instances are generated with the framework in \Cref{sec:design} before
inference. We call one agent attempt on an instance a \emph{run}.

\paragraph{Models and Runs.}
We evaluate Qwen3.8-27B, GPT-OSS-120B, and Gemma-4-31B-it with temperature \(1\), top-\(p=0.95\),
and 16,384 output tokens per turn (32,768 for continuation turns). \Cref{app:experimental-setup}
lists context limits and reasoning settings. The prompt names the target and lists all task
statements and dependencies. Besides the tools from \Cref{sec:design}, the agent has a calculator tool
for arithmetic. In our evaluation, the agent can make at most five \texttt{get\_input} requests per node to limit guessing.
It can correct a report on a later request, but earlier requests still count toward this
limit. The harness rejects malformed submissions and lets the agent retry. It also rejects submissions before the input for the target task is obtained, which GPT-OSS-120B often attempted. After malformed tool calls,
missing submissions, or truncated responses, it sends a limited number of continuation prompts.
These prompts never correct a wrong value reported to \texttt{get\_input}. A run that ends without
an accepted target submission, times out, or exceeds the context limit is incorrect.

\paragraph{Answer Collection.}
Only an accepted \texttt{submit\_answers} call supplies the graded target answer. To find where a
run first goes wrong, we extract answers for every node. We take it from the
\texttt{get\_input} calls of the node's children, since these contain the value the agent passed
on. If no child call contains it, we parse it from the agent's reasoning and response text.

\paragraph{Topologies.}
\Cref{fig:topologies} shows the chain, braid, and fan-out/fan-in topologies, each with 18 nodes.
In each, node 1 is the only root, node 18 is the target, and every other node is an ancestor of the target. The prompt lists tasks
in numerical order, and every parent has a smaller number than its children. The \emph{shuffled
braid} keeps the braid's problems and edges but renumbers the tasks, so some prerequisites are
listed after the tasks that depend on them.
For branch pruning (\Cref{sec:exp-pruning}), we keep all 18 tasks and their inputs fixed and
remove edge~1 in the braid panel of \Cref{fig:topologies}, leaving eight tasks unnecessary for the target but
sharing a root with the required branch. Removing edge~2 as well disconnects the unnecessary
branch completely.

\begin{figure}[t]
    \centering
    \includegraphics[width=\linewidth]{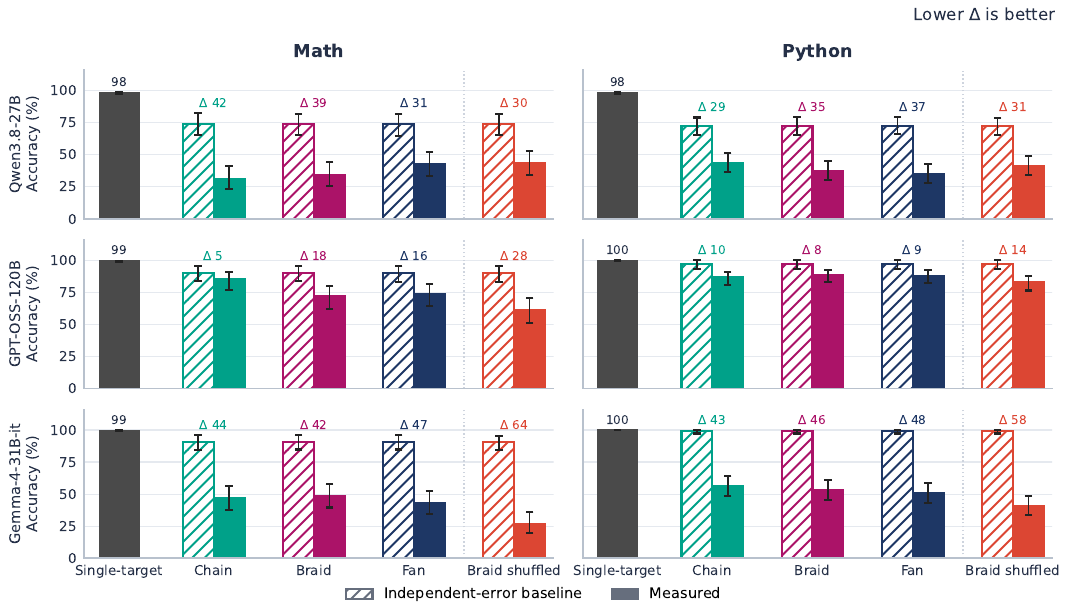}
    \caption{Accuracy on individual tasks and on targets in instances with 18 nodes. Hatched bars
    show the independent-error baseline, and adjacent solid bars show measured target accuracy.
    \(\Delta\) labels give the additional loss in percentage points. Whiskers are 95\%
    bootstrap intervals for baselines and Wilson intervals for measurements. Chain, braid, and fan
    differ in their edges at a fixed number of tasks. The shuffled braid reassigns the braid's
    task numbers without changing its problems or dependencies.}
    \label{fig:gap-results}
\end{figure}

\paragraph{Metrics.}
We report target accuracy with Wilson 95\% intervals. Integers must match exactly, lists are
compared element by element, and floating point values must match within \(10^{-3}\). HotpotQA
uses the official joint exact match, which requires both the answer and the supporting facts to
match.
If errors at different nodes were independent, an instance would succeed only when the target and
all its ancestors succeed. For an instance \(G\) with target \(t\) and ancestors
\(\operatorname{Anc}(t)\), this gives the \emph{independent-error baseline}
$
    \hat q(G,t)=\prod_{i\in\operatorname{Anc}(t)\cup\{t\}}
    \hat p_i^{\mathrm{single}},
$
where \(\hat p_i^{\mathrm{single}}\) is the accuracy on the problem at node \(i\) over repeated
runs in which it is the only node. We average \(\hat q\) over the instances used to measure target
accuracy \(\hat a\) and report the \emph{additional loss} \(\hat q-\hat a\). We bootstrap the
single-node runs to estimate uncertainty in \(\hat q\).
For branch pruning, we measure the fraction of runs with correct answers for all seven
non-leaf tasks on the unnecessary branch. We average the math and Python rates.
To compute 95\% confidence intervals, we resample target tasks within each category,
treating each target's three graph instances as one group.

\section{Experiments}

\label{sec:experiments}
We first run every task alone and use the results to choose tasks for the larger instances. We then
measure how far accuracy on 18-node math-only and Python-only instances falls below the
independent-error baseline, how topology and task order change it, and whether agents skip
branches unnecessary for the target. Finally, we replace half
of the tasks in the math-only instances with tasks from other categories to study the performance on mixed domains.

\subsection{Individual Tasks}
\label{sec:exp-single}
With seed \(0\), we generate eight inputs for each of the 40 math operators, 80 Python tracing
variants, and 40 RLVE objectives, and add 200 questions from HotpotQA's distractor setting. Each of the 1,480 instances
has one node and is run once per model. All models solve more than 91\% of math, Python, and RLVE
tasks (\Cref{tab:single-baselines}). HotpotQA accuracy is 30\% to 40\% under joint exact match,
which also requires the supporting facts. We use these results only to choose tasks. The baselines in
later sections come from separate repeated runs.

\begin{table}[tbp]
    \centering
    \footnotesize
    \caption{Accuracy on instances with one node (\%).}
    \label{tab:single-baselines}
    \begin{tabular}{lrrrr}
        \toprule
        Model & Math & Python & RLVE & HotpotQA \\
        \midrule
        Qwen3.8-27B & 97.2 & 93.3 & 91.9 & 40.0 \\
        GPT-OSS-120B & 97.5 & 98.4 & 94.1 & 30.0 \\
        Gemma-4-31B-it & 98.4 & 96.9 & 93.4 & 32.5 \\
        \bottomrule
    \end{tabular}
\end{table}

\subsection{Accuracy Loss in Workflows}
\label{sec:exp-gap}
For each model, we keep the math operators and Python variants that it solved on all eight
inputs. For each, we fix the first input and require that its alternative-input bank has
distinct answers. We make this selection before any run on a larger instance. For GPT-OSS-120B, we
also drop tasks whose mean response exceeds 12,000 tokens, so that instances fit its context. This
leaves 33 math and 55 Python tasks for Qwen, 32 and 53 for GPT-OSS, and 37 and 52 for Gemma. We
run each math task alone 50 times, and each Python task 45 times for Qwen and 30 times for the
other models. For each target task and topology, we generate three instances whose other 17 nodes
are different tasks of the same category. This gives 99 math and 165 Python instances per topology
for Qwen, 96 and 159 for GPT-OSS, and 111 and 156 for Gemma. All topologies use the same
problems, so they share one baseline. The task sets differ between models; we compare each
model only with its own baseline.

Accuracy is below the baseline in every condition (\Cref{fig:gap-results},
\Cref{tab:prediction-gap}). On Qwen's math chains, the baseline is 73.7\% but accuracy is 31.3\%,
a loss of 42.4 points. Among chains and braids, Gemma's Python braid loses the most, 45.9 points
(99.1\% versus 53.2\%). Without shuffling, GPT-OSS loses 4.5 to 18.1 points in math and 8.2 to 10.1 points in Python.
Only the GPT-OSS math chain has an interval that includes zero. The chain and braid results match a
constant drop per node of 2.8 to 4.6 points for Qwen, 3.1 to 3.6 for Gemma, and 0.3 to 1.3 for
GPT-OSS.

\begin{figure}[tbp]
    \centering
    \includegraphics[width=\linewidth]{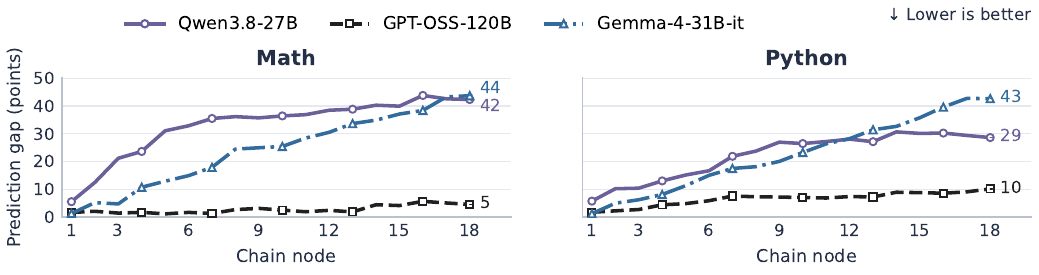}
    \caption{Additional loss along chains with 18 nodes. Each line shows the independent-error
    baseline minus measured correctness of the answer collected at that node, in percentage points.
    Math and Python share the same vertical scale; line-end labels give the target-node losses.}
    \label{fig:chain-depth}
\end{figure}

\Cref{fig:chain-depth} shows additional loss at every node of the chain. Qwen is below its baseline at
the root, while Gemma falls behind at later nodes. Outside the shuffled braid, 78\% to 100\% of
failed instances first go wrong at a node that received its intended input. On Qwen chains, accuracy
at the root, which has no parents, drops from 98.4\% alone to 92.9\% in math and from 97.9\% to
92.1\% in Python. Most failures therefore begin with a wrong solution rather than
a wrong report of a parent answer. The truncation is negligible: only 4 of Qwen's 1,056 runs exceeded the context limit.

\subsection{Topology and Task Order}
\label{sec:exp-topology}
We compute differences between topologies separately for math and Python and combine them with
inverse variance weights, reporting estimates \(\pm\) one standard error. Compared with the chain, the braid changes target accuracy by
$-2.5\pm4.2$ points for Qwen, $-2.5\pm3.1$ for GPT-OSS, and $-1.1\pm4.3$ for Gemma. The
fan-out/fan-in topology changes it by $-1.0\pm4.2$, $-2.5\pm3.1$, and $-4.9\pm4.3$. We find no
consistent effect of the edges on target accuracy, but small effects remain possible.
Compared with the braid, the shuffled braid changes target accuracy by $+5.7\pm4.2$ points for
Qwen, $-6.8\pm3.4$ for GPT-OSS, and $-16.3\pm4.2$ for Gemma. Gemma loses accuracy in both
categories. In math, 36\% of its failures in the shuffled braid begin at a node for which it never
called \texttt{get\_input}, compared with 7\% in the braid.

\subsection{Pruning Unnecessary Branches}
\label{sec:exp-pruning}
We test whether agents skip unnecessary tasks using the two braid variants in \Cref{sec:methodology}.
All three models sometimes solve the disconnected branch (\Cref{tab:branch-pruning}).
GPT-OSS produces correct answers for all seven checked tasks in 83.2\% of runs, compared with
87.2\% when the branch is required. Qwen and Gemma do so in 20.4\% and 18.3\% of
disconnected-branch runs, respectively. Although the branch is unnecessary for the target,
GPT-OSS requests inputs for at least one of its tasks in 98.4\% of runs,
Qwen in 71.9\%, and Gemma in 45.5\%.

\begin{table}[tbp]
    \centering
    \small
    \caption{Fraction of runs with correct answers for all seven verifiable branch tasks (\%).
    The branch is required in the full braid and unnecessary in the two pruned variants.
    Brackets give 95\% confidence intervals.}
    \label{tab:branch-pruning}
    \begin{tabular}{lrrl}
        \toprule
        Model & Full braid & Shared root & Disconnected [95\% CI] \\
        \midrule
        Gemma-4-31B-it & 74.7 & 27.4 & 18.3 [13.6, 23.3] \\
        GPT-OSS-120B & 87.2 & 82.7 & 83.2 [78.5, 87.8] \\
        Qwen3.8-27B & 59.3 & 37.6 & 20.4 [15.5, 25.8] \\
        \bottomrule
    \end{tabular}
\end{table}

\subsection{Instances with Mixed Categories}
\label{sec:exp-mixed}
For each math-only instance in the chain, braid, and fan-out/fan-in topologies, we build a mixed instance
with the same edges, target, and eight other math nodes. The remaining nine nodes are replaced by
five Python, three RLVE, and one HotpotQA task. The RLVE and HotpotQA tasks come from a set of
five RLVE tasks and three HotpotQA questions (\Cref{app:experimental-setup}), each run alone 30
times. This gives 99, 96, and 111 pairs per topology for Qwen, GPT-OSS, and Gemma. We report the
difference in target accuracy between mixed and math-only instances, both directly and after
subtracting the difference in their baselines. The second value removes the effect of the
replacement tasks' difficulty. We compute 95\% intervals by resampling target tasks. For the
adjusted difference, each bootstrap sample also draws the accuracy of each task from its Jeffreys
Beta\((x+\tfrac{1}{2}, n-x+\tfrac{1}{2})\) posterior, where \(x\) of \(n\) single-node runs are
correct.

\begin{figure}[t]
    \centering
    \includegraphics[width=\linewidth]{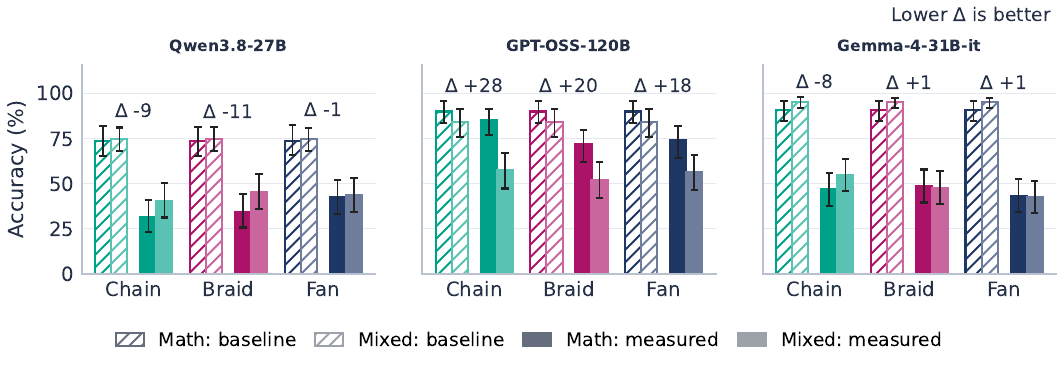}
    \caption{Instances with only math tasks and their paired instances with mixed categories.
    Colors identify topologies as in \Cref{fig:gap-results}; lighter shades denote mixed instances.
    Within each topology, the two hatched bars show the independent-error baselines, followed by
    the two solid bars showing measured accuracy. \(\Delta\) is measured math-only accuracy minus
    mixed accuracy, in percentage points; lower is better. Whiskers are 95\% intervals.}
    \label{fig:mixed-effects}
\end{figure}

The effect of mixing categories depends on the model (\Cref{fig:mixed-effects},
\Cref{tab:mixed-gap}). GPT-OSS loses 21.9 points directly ($[-30.6,-13.2]$) and 16.0 points after
adjustment ($[-25.3,-0.5]$). Qwen gains 7.1 points directly ($[+0.7,+13.5]$) and 6.0 after
adjustment ($[-1.5,+18.3]$). Gemma changes by $+2.1$ ($[-5.7,+10.5]$) and $-2.3$ points
($[-7.3,+13.5]$). The RLVE and HotpotQA tasks were almost always solved alone but were solved
less often inside mixed instances. No run on a mixed instance exceeded the context limit.

\section{Related Work}
\label{sec:background}

\noindent\textbf{Context and long-horizon degradation.}\enspace
Long-context benchmarks show that usable context is limited by distractors, lexical mismatch,
and evidence position~\citep{hsieh2024ruler,bai2024longbench,kuratov2024babilong,modarressi2025nolima,liu2024lostinthemiddle}.
Multi-turn, memory, and agent benchmarks test state across interactions and growing
contexts~\citep{bai2024mtbench101,sirdeshmukh2025multichallenge,maharana2024locomo,wu2024longmemeval,lee2026longmint,zeng2026locabench}.
Accuracy also drops when instructions are split across turns~\citep{laban2025lostinconversation}
and when simple steps are executed over long horizons, in part because models condition on their
own earlier errors~\citep{sinha2026illusion}.
These works vary context or interaction length for one task type.
\benchmarkname{} connects heterogeneous verifiable problems and runs each one alone and inside
workflows, separating the loss caused by the workflow from the difficulty of each problem.

\noindent\textbf{Composing problems for evaluation.}\enspace
Compositional GSM substitutes the answer of one GSM8K problem into a second and finds accuracy
below the product of the two isolated accuracies~\citep{hosseini2024compositionalgsm}; we extend
this baseline to graphs of 18 problems.
Relay-Bench joins subproblems from several domains into one prompt~\citep{swayne2026relaybench},
MuSiQue composes single-hop questions into multi-hop ones~\citep{trivedi2022musique}, REST and SIFo
present several problems or sequential instructions at once~\citep{pan2025rest,chen2024sifo}, and
Robust Reasoning Benchmark finds decay over independent AIME problems solved in
sequence~\citep{golikov2026robustreasoning}.
These benchmarks write dependencies into the problem text, so each combination is authored by
hand or requires compatible formats, and the structure is fixed.
\benchmarkname{} leaves each problem unchanged, checks each reported answer when the agent
requests the next input, and varies topology and task order while the problems stay fixed.
GSM-\(\infty\), DyVal, and Faith and Fate instead build a single problem from a graph of
quantities, premises, or computation steps~\citep{zhou2025gsminfty,zhu2024dyval,dziri2023faith},
whereas our edges connect separate problems.
In agent benchmarks such as NESTFUL, GAIA, and \(\tau\)-bench, one tool output can become the
argument of a later call~\citep{basu2025nestful,mialon2023gaia,yao2024taubench}, but steps are not
measured as standalone problems.

\noindent\textbf{Composing environments for training.}\enspace
RACES fuses verifiable environments whose output and input types match, using composition
operators~\citep{xiang2026races}.
Composition-RL writes an upstream answer into a downstream prompt~\citep{xu2026compositionrl},
and h1 chains GSM8K problems for curriculum reinforcement learning~\citep{motwani2025h1}.
These methods require compatible formats or conversions between them, and a conversion can map a
wrong upstream answer to the correct downstream input (\Cref{sec:adapters}).
\benchmarkname{} instead checks parent answers to select inputs with distinct verified answers.

\section{Limitations and Future Work}
\label{sec:future-work}

We intentionally selected tasks that each model solves reliably in isolation, so that the
independent-error baseline stays high and any additional loss is attributable to the workflow.
Future work could vary what the context contains, such as the
number of listed tasks or the retained reasoning from earlier tasks, to identify which part
causes the additional loss.
Context overflow was rare in our experiments. Instances that exceed the context limit, or agents with memory
and summarization tools, would test whether topology matters once the agent must choose what to
keep. The same verified intermediate answers can also provide rewards for each node when
training with the RLVR interface.

\section{Conclusion}
\label{sec:conclusion}

\benchmarkname{} composes problems with verifiable answers into workflows without matching types
or writing adapters, so tasks from different categories can be combined freely.
In instances with 18 nodes, all three models lose accuracy beyond compounded single-task errors,
by 36 to 45 points on average for Qwen and Gemma and 11 for GPT-OSS. With all nodes required,
topology has no consistent effect on target accuracy, while shuffled task order and mixed
categories hurt some models. Agents also perform unnecessary work on disconnected branches.
Verified intermediate answers show where each failure begins.
We hope \benchmarkname{} helps evaluate and train agents that stay reliable as their tasks become more compositional.

\subsection*{Reproducibility Statement}
We will release the \benchmarkname{} code and all generated datasets upon acceptance.
\Cref{sec:design} describes how instances are built from a topology specification, and explains
how problems are assigned to nodes and how inputs are chosen so that each child input has a
distinct correct answer. All instances are generated with fixed seeds before inference. The task
categories and their sources are listed in \Cref{sec:design}. \Cref{sec:methodology} describes
the models, sampling parameters, harness behavior (retries, continuation prompts, and failure
handling), answer collection, the topologies, and the metrics, including the
independent-error baseline and its bootstrap. All three models have open weights.
\Cref{app:experimental-setup} gives the remaining inference settings (context budgets,
reasoning effort, top-\(k\), and serving setup). It also lists the task-selection procedure,
the IDs of the selected tasks for the mixed instances, and the held-out calibration counts.
\Cref{app:detailed-results} reports detailed benchmark results.

\subsection*{AI use statement}

In this work, we used generative AI tools for proofreading and polishing the writing, including rewriting full paragraphs, writing and reviewing code, rescheduling failing experiments, verifying that we did not miss prior works, and reviewing statistical practices and the correctness of conclusions.
We have not used generative AI tools for writing full sections from scratch, writing the critical code paths, research ideation, or generating datasets.
Math proofs are not applicable to this work.
We have reviewed all AI-assisted work. We
checked references and edited the writing, and all LLM-generated code was verified and tested for correctness. We take responsibility for the final content of this work,
including text, claims, or artifacts produced with the aid of generative AI.

\ificlrfinal
\section*{Acknowledgments}
This work was supported in part by
the \href{https://alliancecan.ca}{Digital Research Alliance of Canada}
and the Vector Institute.

\fi

\bibliography{arbigraph_bibliography_verified}

@inproceedings{bai2024longbench,
  title = "{L}ong{B}ench: A Bilingual, Multitask Benchmark for Long Context Understanding",
  author = "Bai, Yushi and Lv, Xin and Zhang, Jiajie and Lyu, Hongchang and Tang, Jiankai and Huang, Zhidian and Du, Zhengxiao and Liu, Xiao and Zeng, Aohan and Hou, Lei and Dong, Yuxiao and Tang, Jie and Li, Juanzi",
  booktitle = "Proc. of the 62nd Annual Meeting of the Association for Computational Linguistics (Volume 1: Long Papers)",
  month = aug,
  year = "2024",
  url = "https://aclanthology.org/2024.acl-long.172/",
  doi = "10.18653/v1/2024.acl-long.172",
  pages = "3119--3137",
}

@inproceedings{bai2024mtbench101,
  title = "{MT}-Bench-101: A Fine-Grained Benchmark for Evaluating Large Language Models in Multi-Turn Dialogues",
  author = "Bai, Ge and Liu, Jie and Bu, Xingyuan and He, Yancheng and Liu, Jiaheng and Zhou, Zhanhui and Lin, Zhuoran and Su, Wenbo and Ge, Tiezheng and Zheng, Bo and Ouyang, Wanli",
  booktitle = "Proc. of the 62nd Annual Meeting of the Association for Computational Linguistics (Volume 1: Long Papers)",
  month = aug,
  year = "2024",
  url = "https://aclanthology.org/2024.acl-long.401/",
  doi = "10.18653/v1/2024.acl-long.401",
  pages = "7421--7454",
}

@inproceedings{hsieh2024ruler,
  title = {{RULER}: What's the Real Context Size of Your Long-Context Language Models?},
  author = {Cheng-Ping Hsieh and Simeng Sun and Samuel Kriman and Shantanu Acharya and Dima Rekesh and Fei Jia and Yang Zhang and Boris Ginsburg},
  booktitle={Proc. of the 1st Conference on Language Modeling (COLM)},
  year = {2024},
  url = {https://arxiv.org/abs/2404.06654},
}

@inproceedings{kuratov2024babilong,
  title = {{BABILong}: Testing the Limits of {LLM}s with Long Context Reasoning-in-a-Haystack},
  author = {Yuri Kuratov and Aydar Bulatov and Petr Anokhin and Ivan Rodkin and Dmitry Sorokin and Artyom Sorokin and Mikhail Burtsev},
  booktitle	= {Proc. International Conference on Neural Information
		  Processing Systems (NeurIPS)},
  pages = {106519--106554},
  year = {2024},
  doi = {10.52202/079017-3381}
}

@misc{lee2026longmint,
  title = {{LongMINT}: Evaluating Memory under Multi-Target Interference in Long-Horizon Agent Systems},
  author = {Hyunji Lee and Justin Chih-Yao Chen and Joykirat Singh and Zaid Khan and Elias Stengel-Eskin and Mohit Bansal},
  year = {2026},
  eprint = {2605.18565},
  archivePrefix = {arXiv},
  primaryClass = {cs.CL},
  url = {https://arxiv.org/abs/2605.18565},
  note = {arXiv preprint arXiv:2605.18565}
}

@article{liu2024lostinthemiddle,
  title = {Lost in the Middle: How Language Models Use Long Contexts},
  author = {Liu, Nelson F. and Lin, Kevin and Hewitt, John and Paranjape, Ashwin and Bevilacqua, Michele and Petroni, Fabio and  Liang, Percy},
  journal = {Transactions of the Association for Computational Linguistics},
  volume = {12},
  pages = {157--173},
  year = {2024},
  doi = {10.1162/tacl_a_00638},
  url = {https://doi.org/10.1162/tacl_a_00638},
}

@inproceedings{maharana2024locomo,
    title = "Evaluating Very Long-Term Conversational Memory of {LLM} Agents",
    author = "Maharana, Adyasha  and
      Lee, Dong-Ho  and
      Tulyakov, Sergey  and
      Bansal, Mohit  and
      Barbieri, Francesco  and
      Fang, Yuwei",
    booktitle = "Proc. of the 62nd Annual Meeting of the Association for Computational Linguistics (Volume 1: Long Papers)",
    month = aug,
    year = "2024",
    url = "https://aclanthology.org/2024.acl-long.747/",
    doi = "10.18653/v1/2024.acl-long.747",
    pages = "13851--13870",
}

@inproceedings{mialon2023gaia,
 author = {Mialon, Gr\'{e}goire and Fourrier, Cl\'{e}mentine and Wolf, Thomas and LeCun, Yann and Scialom, Thomas},
 booktitle = {Proc. of the International Conference on Learning Representations (ICLR)},
 pages = {9025--9049},
 title = {{GAIA}: a benchmark for General AI Assistants},
 url = {https://proceedings.iclr.cc/paper_files/paper/2024/file/25ae35b5b1738d80f1f03a8713e405ec-Paper-Conference.pdf},
 year = {2024}
}

@inproceedings{modarressi2025nolima,
  title = {{NoLiMa}: Long-Context Evaluation Beyond Literal Matching},
  author = {Modarressi, Ali and Deilamsalehy, Hanieh and Dernoncourt, Franck and Bui, Trung and Rossi, Ryan A. and Yoon, Seunghyun and Sch{\"u}tze, Hinrich},
  booktitle = {Proc. of the 42nd International Conference on Machine Learning (ICML)},
  year = {2025}
}

@inproceedings{sirdeshmukh2025multichallenge,
    title = "{M}ulti{C}hallenge: A Realistic Multi-Turn Conversation Evaluation Benchmark Challenging to Frontier {LLM}s",
    author = "Deshpande, Kaustubh  and
      Sirdeshmukh, Ved  and
      Mols, Johannes Baptist  and
      Jin, Lifeng  and
      Hernandez-Cardona, Ed-Yeremai  and
      Lee, Dean  and
      Kritz, Jeremy  and
      Primack, Willow E.  and
      Yue, Summer  and
      Xing, Chen",
    booktitle = "Findings of the Association for Computational Linguistics: ACL 2025",
    month = jul,
    year = "2025",
    url = "https://aclanthology.org/2025.findings-acl.958/",
    doi = "10.18653/v1/2025.findings-acl.958",
    pages = "18632--18702",
}

@inproceedings{wu2024longmemeval,
  title = {{LongMemEval}: Benchmarking Chat Assistants on Long-Term Interactive Memory},
  author = {Di Wu and Hongwei Wang and Wenhao Yu and Yuwei Zhang and Kai-Wei Chang and Dong Yu},
  booktitle = {Proc. of the International Conference on Learning Representations (ICLR)},
  year = {2025}
}

@misc{xiang2026races,
  title = {Verifiable Environments Are {LEGO} Bricks: Recursive Composition for Reasoning Generalization},
  author = {Hao Xiang and Qiaoyu Tang and Le Yu and Yaojie Lu and Xianpei Han and Ben He and Le Sun and Bowen Yu and Peng Wang and Hongyu Lin and Dayiheng Liu},
  year = {2026},
  eprint = {2606.12373},
  archivePrefix = {arXiv},
  primaryClass = {cs.AI},
  url = {https://arxiv.org/abs/2606.12373},
  note = {arXiv preprint arXiv:2606.12373}
}

@inproceedings{yao2024taubench,
  title = {{$\tau$-bench}: A Benchmark for Tool-Agent-User Interaction in Real-World Domains},
  author = {Shunyu Yao and Noah Shinn and Pedram Razavi and Karthik Narasimhan},
  booktitle = {Proc. of the International Conference on Learning Representations (ICLR)},
  year = {2025},
  url = {https://openreview.net/forum?id=roNSXZpUDN},
}

@inproceedings{zeng2026locabench,
  title = {{LOCA-bench}: Benchmarking Language Agents Under Controllable and Extreme Context Growth},
  author = {Weihao Zeng and Yuzhen Huang and Junxian He},
  year = {2026},
  booktitle = {Proc. of the International Conference on Machine Learning (ICML)},
  url={https://openreview.net/forum?id=N4HIJq8v95},
}

@inproceedings{pan2025rest,
    title = "{REST}: Stress Testing Large Reasoning Models by Asking Multiple Problems at Once",
    author = "Pan, Zhuoshi  and
      Pei, Qizhi  and
      Li, Yu  and
      Tang, Zinan  and
      Sun, QiYao  and
      Zhao, H. Vicky  and
      He, Conghui  and
      Wu, Lijun",
    booktitle = "Proc. of the 64th Annual Meeting of the {A}ssociation for {C}omputational {L}inguistics (Volume 1: Long Papers)",
    month = jul,
    year = "2026",
    url = "https://aclanthology.org/2026.acl-long.1296/",
    doi = "10.18653/v1/2026.acl-long.1296",
    pages = "28110--28140",
}

@inproceedings{zhou2025gsminfty,
  title={{GSM}-Infinite: How Do your {LLM}s Behave over Infinitely Increasing Reasoning Complexity and Context Length?},
  author={Yang Zhou and Hongyi Liu and Zhuoming Chen and Yuandong Tian and Beidi Chen},
  booktitle={Proc. of the ICML Workshop on Long-Context Foundation Models},
  year={2025},
  url={https://openreview.net/forum?id=FGoZOZzotG}
}

@inproceedings{zhu2024dyval,
  title = {{DyVal}: Dynamic Evaluation of Large Language Models for Reasoning Tasks},
  author = {Kaijie Zhu and Jiaao Chen and Jindong Wang and Neil Zhenqiang Gong and Diyi Yang and Xing Xie},
  booktitle = {Proc. of the International Conference on Learning Representations (ICLR)},
  year = {2024},
  url = {https://openreview.net/forum?id=gjfOL9z5Xr}
}

@inproceedings{chen2024sifo,
  title = "The {SIFo} Benchmark: Investigating the Sequential Instruction Following Ability of Large Language Models",
  author = "Chen, Xinyi and Liao, Baohao and Qi, Jirui and Eustratiadis, Panagiotis and Monz, Christof and Bisazza, Arianna and de Rijke, Maarten",
  booktitle = "Findings of the Association for Computational Linguistics: EMNLP 2024",
  month = nov,
  year = "2024",
  url = "https://aclanthology.org/2024.findings-emnlp.92/",
  doi = "10.18653/v1/2024.findings-emnlp.92",
  pages = "1691--1706",
}

@inproceedings{motwani2025h1,
  title={{h1}: Bootstrapping {LLM}s to Reason over Longer Horizons via Reinforcement Learning},
  author={Ivanova, Alesia and Motwani, Sumeet Ramesh and Cai, Ziyang and Torr, Philip and Islam, Riashat and Shah, Shital and Schroeder de Witt, Christian and London, Charles},
  booktitle={Proc. of the International Conference on Machine Learning (ICML)},
  year={2026},
  url={https://openreview.net/forum?id=3BW15kSPfN}
}

@misc{xu2026compositionrl,
  title = {{Composition-RL}: Compose Your Verifiable Prompts for Reinforcement Learning of Large Language Models},
  author = {Xin Xu and Clive Bai and Kai Yang and Tianhao Chen and Yangkun Chen and Weijie Liu and Hao Chen and Yang Wang and Saiyong Yang and Can Yang},
  year = {2026},
  eprint = {2602.12036},
  archivePrefix = {arXiv},
  primaryClass = {cs.LG},
  url = {https://arxiv.org/abs/2602.12036},
  note = {arXiv preprint arXiv:2602.12036}
}

@misc{golikov2026robustreasoning,
  title = {Robust Reasoning Benchmark},
  author = {Golikov, Pavel and Opryshko, Evgenii and Pekhimenko, Gennady and Jeffrey, Mark C.},
  year = {2026},
  eprint = {2604.08571},
  archivePrefix = {arXiv},
  primaryClass = {cs.LG},
  doi = {10.48550/arXiv.2604.08571},
  url = {https://arxiv.org/abs/2604.08571},
  note = {arXiv preprint arXiv:2604.08571}
}

@inproceedings{zeng2026rlve,
  title = {{RLVE}: Scaling Up Reinforcement Learning for Language Models with Adaptive Verifiable Environments},
  author = {Zeng, Zhiyuan and Ivison, Hamish and Wang, Yiping and Yuan, Lifan and Li, Shuyue Stella and Ye, Zhuorui and Li, Siting and He, Jacqueline and Zhou, Runlong and Chen, Tong and Zhao, Chenyang and Tsvetkov, Yulia and Du, Simon Shaolei and Jaques, Natasha and Peng, Hao and Koh, Pang Wei and Hajishirzi, Hannaneh},
  booktitle = {International Conference on Machine Learning (ICML)},
  year = {2026},
  url = {https://arxiv.org/abs/2511.07317}
}

@inproceedings{yang2018hotpotqa,
  title = {{HotpotQA}: A Dataset for Diverse, Explainable Multi-hop Question Answering},
  author = {Yang, Zhilin and Qi, Peng and Zhang, Saizheng and Bengio, Yoshua and Cohen, William W. and Salakhutdinov, Ruslan and Manning, Christopher D.},
  booktitle = {Proceedings of the 2018 Conference on Empirical Methods in Natural Language Processing},
  pages = {2369--2380},
  year = {2018},
  url = {https://aclanthology.org/D18-1259/}
}

@misc{leetcode_dataset,
      title={LeetCodeDataset: A Temporal Dataset for Robust Evaluation and Efficient Training of Code LLMs}, 
      author={Yunhui Xia and Wei Shen and Yan Wang and Jason Klein Liu and Huifeng Sun and Siyue Wu and Jian Hu and Xiaolong Xu},
      year={2025},
      eprint={2504.14655},
      archivePrefix={arXiv},
      primaryClass={cs.LG},
      url={https://arxiv.org/abs/2504.14655}, 
}

@inproceedings{hagberg2008networkx,
  title = {Exploring Network Structure, Dynamics, and Function using {NetworkX}},
  author = {Hagberg, Aric A. and Schult, Daniel A. and Swart, Pieter J.},
  booktitle = {Proc. of the Python in Science Conference},
  pages = {11--15},
  doi = {10.25080/TCWV9851},
  url = {https://networkx.org/en/},
  month = aug,
  year = {2008}
}

@incollection{ellson2004graphviz,
  title = {Graphviz and Dynagraph -- Static and Dynamic Graph Drawing Tools},
  author = {John Ellson and Emden R. Gansner and Eleftherios Koutsofios and Stephen C. North and Gordon Woodhull},
  booktitle = {Graph Drawing Software},
  series = {Mathematics and Visualization},
  pages = {127--148},
  publisher = {Springer},
  doi = {10.1007/978-3-642-18638-7_6},
  url = {https://graphviz.org/},
  year = {2004}
}

@misc{hosseini2024compositionalgsm,
  title = {Not All {LLM} Reasoners Are Created Equal},
  author = {Hosseini, Arian and Sordoni, Alessandro and Toyama, Daniel and Courville, Aaron and Agarwal, Rishabh},
  year = {2024},
  eprint = {2410.01748},
  archivePrefix = {arXiv},
  url = {https://arxiv.org/abs/2410.01748},
  note = {arXiv preprint arXiv:2410.01748}
}

@inproceedings{sinha2026illusion,
  title = {The Illusion of Diminishing Returns: Measuring Long Horizon Execution in {LLM}s},
  author = {Sinha, Akshit and Arun, Arvindh and Goel, Shashwat and Staab, Steffen and Geiping, Jonas},
  booktitle = {Proc. of the International Conference on Learning Representations (ICLR)},
  year = {2026},
  url = {https://arxiv.org/abs/2509.09677}
}

@misc{laban2025lostinconversation,
  title = {{LLM}s Get Lost In Multi-Turn Conversation},
  author = {Laban, Philippe and Hayashi, Hiroaki and Zhou, Yingbo and Neville, Jennifer},
  year = {2025},
  eprint = {2505.06120},
  archivePrefix = {arXiv},
  url = {https://arxiv.org/abs/2505.06120},
  note = {arXiv preprint arXiv:2505.06120}
}

@inproceedings{dziri2023faith,
  title = {Faith and Fate: Limits of Transformers on Compositionality},
  author = {Dziri, Nouha and Lu, Ximing and Sclar, Melanie and Li, Xiang Lorraine and Jiang, Liwei and Lin, Bill Yuchen and West, Peter and Bhagavatula, Chandra and Le Bras, Ronan and Hwang, Jena D. and Sanyal, Soumya and Welleck, Sean and Ren, Xiang and Ettinger, Allyson and Harchaoui, Zaid and Choi, Yejin},
  booktitle = {Proc. of the International Conference on Neural Information Processing Systems (NeurIPS)},
  year = {2023},
  url = {https://arxiv.org/abs/2305.18654}
}

@article{trivedi2022musique,
  title = {{MuSiQue}: Multihop Questions via Single-hop Question Composition},
  author = {Trivedi, Harsh and Balasubramanian, Niranjan and Khot, Tushar and Sabharwal, Ashish},
  journal = {Transactions of the Association for Computational Linguistics},
  volume = {10},
  pages = {539--554},
  year = {2022},
  url = {https://aclanthology.org/2022.tacl-1.31/}
}

@misc{swayne2026relaybench,
  title = {{Relay-Bench}: Evaluating {LLM}s on Multi-Domain Reasoning Chains},
  author = {Swayne, Liam},
  year = {2026},
  eprint = {2607.18438},
  archivePrefix = {arXiv},
  url = {https://arxiv.org/abs/2607.18438},
  note = {arXiv preprint arXiv:2607.18438}
}

@inproceedings{basu2025nestful,
  title = {{NESTFUL}: A Benchmark for Evaluating {LLM}s on Nested Sequences of {API} Calls},
  author = {Basu, Kinjal and Abdelaziz, Ibrahim and Kate, Kiran and Agarwal, Mayank and Crouse, Maxwell and Rizk, Yara and Bradford, Kelsey and Munawar, Asim and Kumaravel, Sadhana and Goyal, Saurabh and Wang, Xin and Lastras, Luis A. and Kapanipathi, Pavan},
  booktitle = {Proc. of the Conference on Empirical Methods in Natural Language Processing (EMNLP)},
  pages = {33538--33547},
  year = {2025},
  url = {https://aclanthology.org/2025.emnlp-main.1702/}
}
\bibliographystyle{iclr2027_conference}

\appendix
\section{Experimental Setup Details}
\label{app:experimental-setup}

All models were served with vLLM on H100 GPUs.
The workflow-study context budgets were 262k tokens for Qwen3.8-27B and Gemma-4-31B-it,
and 131k for GPT-OSS-120B. Qwen used reasoning effort xhigh and top-\(k=20\);
GPT-OSS used effort high, top-\(k=20\), and locally rendered Harmony prompts;
Gemma used thinking mode and top-\(k=64\). Each model's repeated single-node runs used the
same inference settings as its corresponding runs on instances. Results for all instances count context overflows,
malformed tool calls, and missing accepted target submissions as incorrect.
Problems in the instances with 18 nodes use sample index zero from the broad one-node datasets,
chosen before any results from the instances were observed.
The candidate task pool for mixed instances contained 24 questions from HotpotQA's distractor setting and 12 RLVE tasks.
Among candidates that passed eight of eight single-target prescreen runs on all three models,
we chose the three questions with the longest generated responses and five RLVE tasks
nearest the math/Python output-length median after a 1,800-token minimum. The resulting
IDs were HotpotQA 68, 25, and 184 and RLVE 13, 31, 38, 21, and 11.
Held-out calibration then used
30 fresh repeats per task and model: 90 HotpotQA and 150 RLVE runs per model. The resulting
HotpotQA/RLVE correct counts were 88/90 and 148/150 for Qwen, 87/90 and 146/150 for GPT-OSS,
and 90/90 and 150/150 for Gemma.

\section{Detailed Benchmark Results}
\label{app:detailed-results}

\Cref{tab:prediction-gap} reports all four conditions with one task category. The
``Baseline'' column gives the independent-error baseline, computed from repeated single-node accuracy of the same problems,
not from the one-node results in \Cref{tab:single-baselines}. Each model's three structural topologies
and shuffled braid use the same target problems and single-node runs.

\begin{table}[htbp]
    \centering
    \small
    \setlength{\tabcolsep}{4pt}
    \caption{Single-node accuracy and target accuracy in instances with 18 nodes (\%). ``Fan'' is fan-out/fan-in;
    ``Shuffle'' reassigns braid task numbers while keeping the problems and dependencies fixed.
    Except in the shuffled braid, tasks are listed in numerical order. Each condition has 99/165
    Qwen, 96/159 GPT-OSS, and 111/156 Gemma math/Python instances.}
    \label{tab:prediction-gap}
    \begin{tabular}{llrrrrrr}
        \toprule
        Model & Category & Single-target & Baseline & Chain & Braid & Fan & Shuffle \\
        \midrule
        Qwen3.8-27B & Math   & 98.4 & 73.7 & 31.3 & 34.3 & 42.4 & 43.4 \\
                      & Python & 98.3 & 72.3 & 43.6 & 37.6 & 35.2 & 41.2 \\
        GPT-OSS-120B & Math   & 99.4 & 90.0 & 85.4 & 71.9 & 74.0 & 61.5 \\
                      & Python & 99.8 & 96.9 & 86.8 & 88.7 & 88.1 & 83.0 \\
        Gemma-4-31B-it & Math   & 99.5 & 90.6 & 46.8 & 48.6 & 43.2 & 27.0 \\
                       & Python & 99.9 & 99.1 & 56.4 & 53.2 & 50.6 & 41.0 \\
        \bottomrule
    \end{tabular}
\end{table}

\Cref{tab:mixed-gap} gives the per-topology paired comparison. Its adjusted
difference is the measured mixed-minus-pure difference less the change in the
independent-error baseline. The paired instances retain the same math target and eight
other math nodes; only the remaining nine nodes change.

\begin{table}[htbp]
    \centering
    \small
    \caption{Paired target accuracy for instances with only math tasks versus mixed categories (\%). Differences
    are mixed minus pure in percentage points and are computed before rounding.
    The change in the independent-error baseline from pure to mixed is
    $+1.1$ points for Qwen, $-5.9$ for GPT-OSS, and $+4.4$ for Gemma.}
    \label{tab:mixed-gap}
    \begin{tabular}{llrrrr}
        \toprule
        Model & Topology & Pure & Mixed & Direct & Adjusted \\
        \midrule
        Qwen3.8-27B & Chain & 31.3 & 40.4 & $+9.1$ & $+8.0$ \\
                     & Braid & 34.3 & 45.5 & $+11.1$ & $+10.0$ \\
                     & Fan   & 42.4 & 43.4 & $+1.0$ & $-0.1$ \\
        GPT-OSS-120B & Chain & 85.4 & 57.3 & $-28.1$ & $-22.2$ \\
                     & Braid & 71.9 & 52.1 & $-19.8$ & $-13.9$ \\
                     & Fan   & 74.0 & 56.2 & $-17.7$ & $-11.8$ \\
        Gemma-4-31B-it & Chain & 46.8 & 55.0 & $+8.1$ & $+3.7$ \\
                       & Braid & 48.6 & 47.7 & $-0.9$ & $-5.3$ \\
                       & Fan   & 43.2 & 42.3 & $-0.9$ & $-5.3$ \\
        \bottomrule
    \end{tabular}
\end{table}

\end{document}